\documentclass[sigconf]{acmart}
\AtBeginDocument{%
  }

\renewcommand\footnotetextcopyrightpermission[1]{} %

\usepackage{hyperref}
\usepackage{multirow}
\usepackage{makecell}
\usepackage{tabu}
\usepackage{caption}
\usepackage{subcaption}
\usepackage[linesnumbered,ruled,vlined]{algorithm2e}
\usepackage{cleveref}

\begin{document}
\fancyhf{} %
\pagestyle{empty} %

\title{DiffuMural: Restoring Dunhuang Murals with Multi-scale Diffusion}

\author{Puyu Han$^1$\quad Jiaju Kang$^2$\quad Yuhang Pan$^3$\quad Erting Pan$^4$\quad Zeyu Zhang$^5$\\
Qunchao Jin$^6$\quad Juntao Jiang$^7$\quad Zhichen Liu$^1$\quad Luqi Gong$^8$\\
\vspace{0.2cm}
$^1$Southern University of Science and Technology\quad $^2$Beijing Normal University\\
$^3$Hebei Guoyan Science and Technology Center\quad
$^4$Wuhan University\\
$^5$The Australian National University\quad
$^6$AI Geeks\\
$^7$Zhejiang University\quad
$^8$Zhejiang Lab}

\renewcommand{\shortauthors}{Trovato et al.}

\begin{abstract}

Large-scale pre-trained diffusion models have produced excellent results in the field of conditional image generation. However, restoration of ancient murals, as an important downstream task in this field, poses significant challenges to diffusion model-based restoration methods due to its large defective area and scarce training samples. Conditional restoration tasks are more concerned with whether the restored part meets the aesthetic standards of mural restoration in terms of overall style and seam detail, and such metrics for evaluating heuristic image complements are lacking in current research. We therefore propose DiffuMural, a combined Multi-scale convergence and Collaborative Diffusion mechanism with ControlNet and cyclic consistency loss to optimise the matching between the generated images and the conditional control. DiffuMural demonstrates outstanding capabilities in mural restoration, leveraging training data from 23 large-scale Dunhuang murals that exhibit consistent visual aesthetics. The model excels in restoring intricate details, achieving a coherent overall appearance, and addressing the unique challenges posed by incomplete murals lacking factual grounding. Our evaluation framework incorporates four key metrics to quantitatively assess incomplete murals: factual accuracy, textural detail, contextual semantics, and holistic visual coherence. Furthermore, we integrate humanistic value assessments to ensure the restored murals retain their cultural and artistic significance. Extensive experiments validate that our method outperforms state-of-the-art (SOTA) approaches in both qualitative and quantitative metrics.

\end{abstract}



\maketitle

\section{Introduction}
\label{sec:intro}
The Mogao Grottoes in Dunhuang, a World Heritage Site, shelter the largest and most well-preserved ancient murals~\cite{qu2014conservation}. These invaluable masterpieces, as non-renewable resources, have sustained considerable damage due to both natural factors and human activities over a long time. A series of degradations such as cracks, flaking, structural collapse, and fading colors (refer to Fig.~\ref{fig:challenge}) have tarnished their original splendor, marking a profound loss to human civilization~\cite{mccormick2005death}. Faced with the risks of further irreversible damage to the delicate surfaces of the murals and the labor-intensive costs, the cultural heritage community has grown increasingly cautious about relying on hand-coloring methods. In contrast, digital restoration, utilizing digital high-resolution murals as a medium, offers a greater tolerance for error and presents a broader array of solutions for the restoration process~\cite{xu2024comprehensive}.%

Thus, the exploration of effective digital techniques for restoring these murals in high-resolution is of paramount importance for large-scale restoration and the long-term preservation of these cultural treasures~\cite{wang2018understanding, wang2024methods}.

\begin{figure}[t]
\centering
\includegraphics[width=\linewidth]{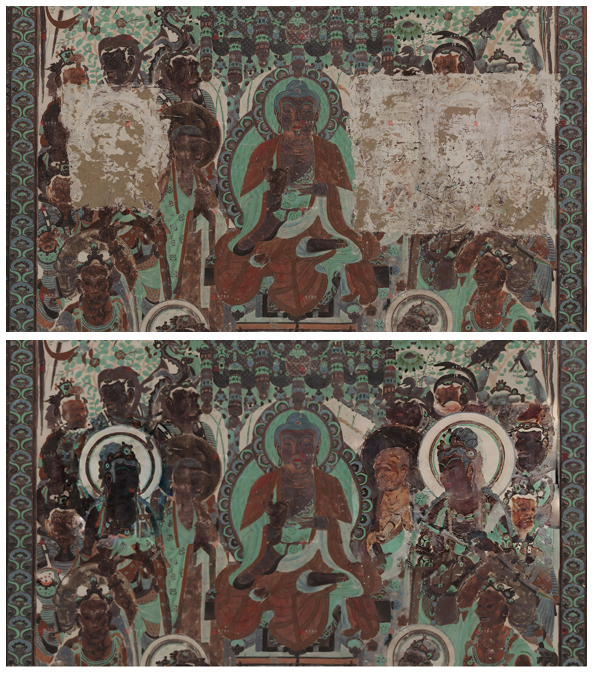}
\caption{Disruption image and the restoration result of a mural in Eastern Wall, Cave 320, Dunhuang with our DiffuMural model.}
\label{fig:restoration}
\end{figure}

Deep learning-based image inpainting and restoration methods have made considerable strides in recent years~\cite{zhang2023image}, %
yet they may encounter significant challenges when applied to the restoration of murals. This is due to the unique nature of mural painting, which involves the creation of smooth lines and evocative colors that differ from those typically found in natural images. As such, inpainting techniques designed for natural images may not be well-suited for restoring the distinct edge structures and intricate texture details inherent in murals. Furthermore, the varying degrees of damage to murals often alter the nature of the restoration task, and nearly all current digital restoration technologies are primarily focused on simpler restoration challenges, such as filling small defects and restoring localized color. %

Notably, heavily damaged murals (refer to Fig.~\ref{fig: di} upper part) may have lost substantial amounts of information, rendering the inpainting process ineffective when simulated and learned through existing rectangular or arbitrary masks. Furthermore, we observed that the damaged areas contain a wealth of vivid and meaningful content, which holds immense potential to provide crucial insights for the restoration process and should not be overlooked. %
Inspired by this, we integrate the spatial dynamics of damaged regions in high-resolution murals and propose a generative framework, DiffuMural, for the task of restoring large areas of the missing mural. 

The main contributions can be summarized below:
\begin{enumerate}

    \item \textbf{Self-Guidance via Damage Contour}: Taking segmentation masks of damaged regions as conditional guidance, ensuring stylistic and textural consistency between the restored and undamaged parts of the mural.
    \item \textbf{Feature Fusion Across Multi-Scale}: Integrating information from both low-resolution and high-resolution layers, balancing global structure and local details to enhance the restoration quality and consistency, even with limited sample data.
    \item \textbf{Co-Diffusion Across Multi-Scale}: Propagating high-confidence information across layers of different resolutions, resulting in a more realistic and seamless integration of the restored areas with the original mural.
    \item \textbf{Humanistic Value Alignment Evaluation}: Evaluating the restoration performance with mural restoration experts, ensuring that the restored sections seamlessly blended with the cultural and historical significance of the original artwork.
\end{enumerate}

\begin{figure}[t]
\centering
\includegraphics[width=\linewidth]{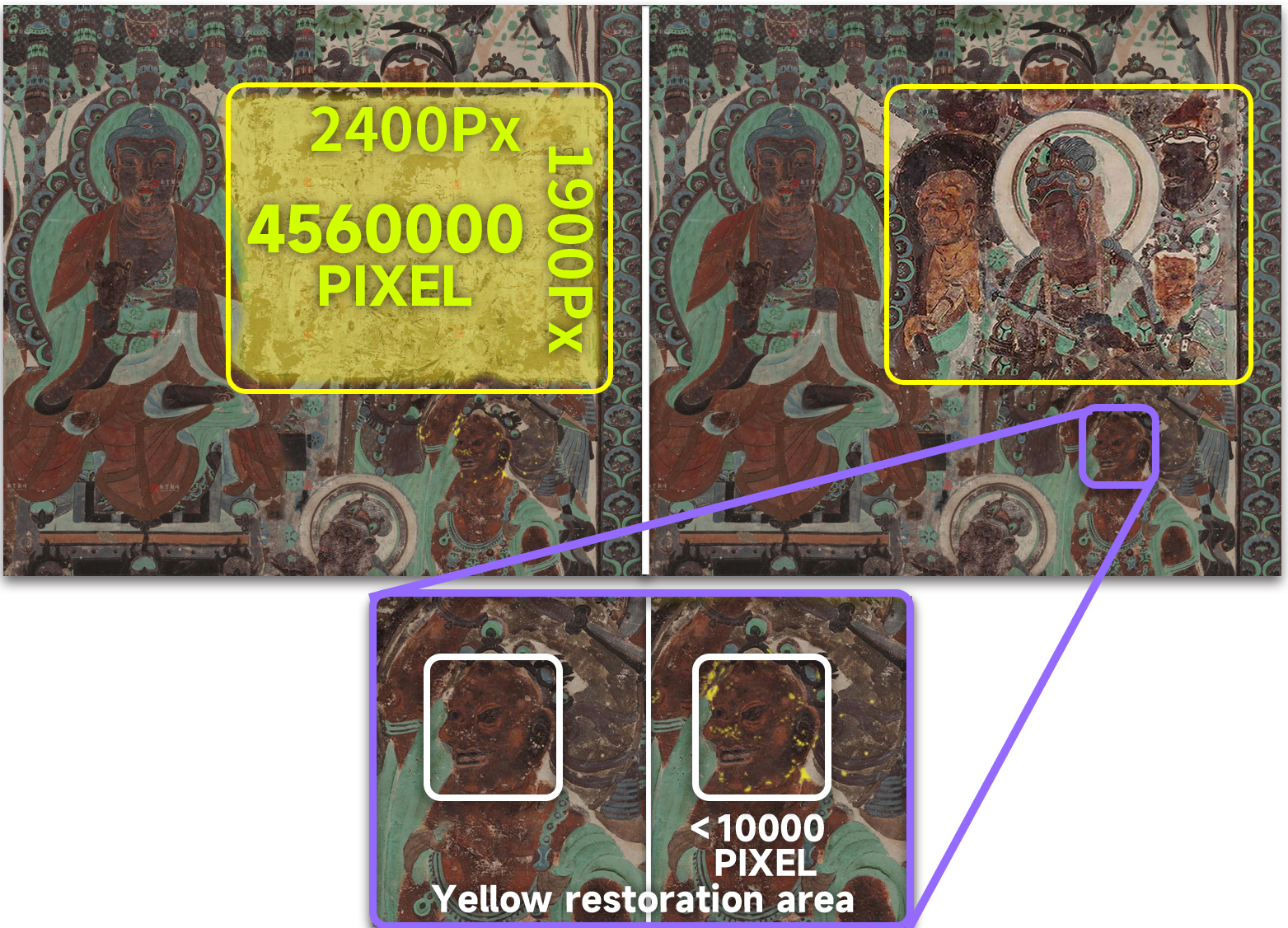}
\caption{The challenge of our restoration task.}
\label{fig:challenge}
\end{figure}

\section{Related work}
\label{sec:related}

\begin{figure}[t]
\centering
\includegraphics[width=\linewidth]{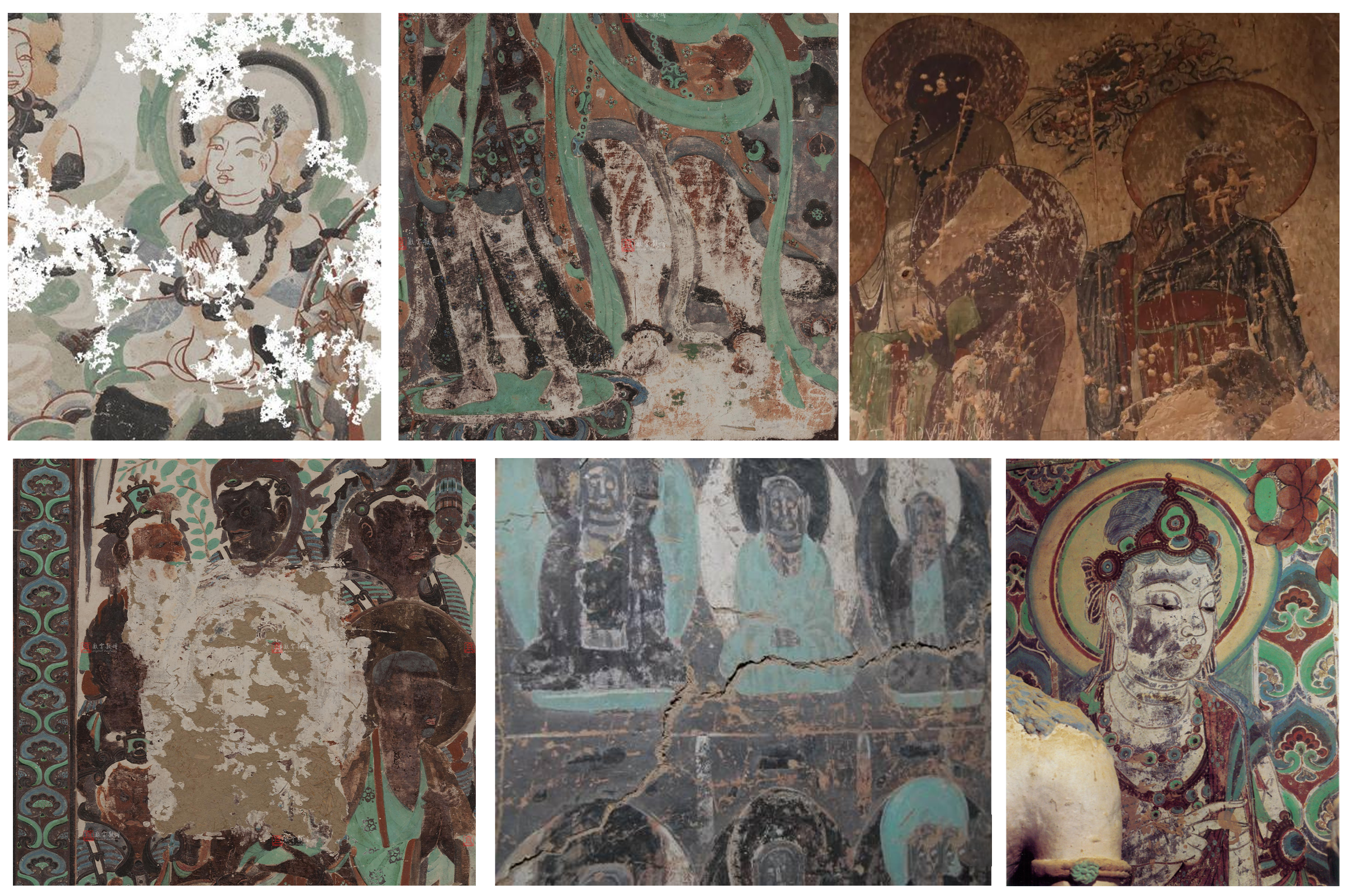}
\caption{Types of mural damages: cracks, flaking, structural collapse, and fading colors}

\label{fig:challenge}
\end{figure}
\textbf{Digital restoration of ancient mural}.
AI-based mural restoration involves using deep learning methods to reconstruct damaged murals while preserving their original style and detail. Key challenges include style fidelity, handling large missing areas, and balancing artistic authenticity with visual accuracy. GAN-based methods \cite{cao2020ancient,wu2021clothgan,wang2021virtual,li2021restoration,cao2021superresolution,ma2022improved,ren2024dunhuang,yan2024image} have demonstrated notable effectiveness in generating visually coherent and stylistically consistent results by leveraging adversarial training to recreate fine details and textures. Diffusion model-based methods \cite{zhang2025fdg,zhang2024motion} have also been widely used in this task \cite{huang2023diffusion,shao2023building,xu2024muraldiff}, showing a strong capability to model complex structures and nuanced artistic patterns even with substantial damage, making them well-suited for the intricate textures often found in ancient murals. More effective design with the generative model can enhance mural restoration by allowing precise reconstruction of details and artistic elements. 

\noindent\textbf{Conditional guidance generation.}
Conditional generation is a type of task for generative models that uses conditional inputs to control the features of generated outputs. This task offers significant flexibility and controllability, allowing the model to produce desired outputs based on varying input conditions, which has important applications in image restoration. The core challenge lies in maintaining high generation quality while ensuring a strong correlation between the condition and the generated content. Attempts focus on introducing conditional information into generative models. VAE-based methods \cite{sohn2015learning,pagnoni2018conditional,harvey2021conditional} have high stability, are easy to train, and can effectively integrate conditional information. GAN-based theories and applications \cite{mirza2014conditional, denton2016semi, isola2017image,lin2018conditional,chrysos2018robust,zhang2018decoupled,chen2019improving, ding2021ccgan} feed conditional information to both the generator and discriminator, allowing the generator to produce data in a specific style based on the given condition. Conditional diffusion-based models \cite{batzolis2021conditional,sinha2021d2c,huang2022fastdiff,zhu2023conditional,chen2023seeing,zhang2023shiftddpms,ni2023conditional,baldassari2024conditional} incorporate conditional information into the generation process, producing outputs under specific conditions by controlling diffusion steps. These models demonstrate powerful generative capabilities in tasks such as text and image generation, particularly excelling in high-resolution and detail-rich generation tasks. ControlNet \cite{zhang2023adding} integrates additional guidance, such as edges, depth, or pose information, allowing the model to generate content with greater alignment to specified details, showing great success. 

\noindent\textbf{Large-scale pre-training.}
Large-scale pre-training has become a cornerstone in modern machine learning, especially in tasks like image generation.  Large-scale pre-training enables models like CLIP ~\cite{radford2021learning} to learn deep semantic relationships between images and text by training on vast amounts of image-text pairs. Generative models like Stable Diffusion ~\cite{rombach2022high} can leverage the mapping provided by CLIP to create high-quality images based on textual descriptions.  Large-scale pre-training has been widely used in image retortion and inpainting \cite{li2022mat, liu2022siamtrans,li2023lsdir,luo2024photo,dudhane2024dynamic,xu2024boosting,wang2024image}. By learning from vast amounts of data, pre-trained models can easily understand the context and structure of images, which helps in tasks like inpainting or repairing damaged images, filling in missing parts of an image with realistic details, drawing from learned knowledge of how objects and scenes should appear, offering enhanced accuracy, creativity, and context-aware solutions for repairing and reconstructing images.

\section{Overview}
The mural restoration task described in this paper was proposed by the Dunhuang Research Institute. In contrast to previous mural restoration tasks, this is an exploratory restoration of a large area lossing mural, which currently lacks adequate evaluation metrics. Consequently, we have introduced quantitative indicators, including colour consistency, texture consistency, edge consistency, structural similarity, and others, as well as a human value assessment system comprising professional mural restorers to evaluate the restoration results.

In section 4, we will give the restoration process and the corresponding theoretical basis for the mural paintings in Dunhuang Cave No. 320, and compare the results with other models to verify the validity of the MuralDiff and to explore the practical application value of the MuralDiff.

\section{Restoration processes and Methodology}
For the restoration of large-area lossing murals, the existing algorithms are not applicable. The main challenges are listed: the inference of large-area missing regions requires the restoration algorithms to have strong contextual understanding and to reasonably mine effective guidance information to assist in the inference. In addition, with limited data samples, we need to consider how to maximise the intake of effective training information to ensure the model generation capability.

For the first challenge, we also propose a conditionally guided mural controllable generation framework, which enhances the controllability of mural generation by extracting the hidden and fuzzy contours of the damaged regions and introduces the mural spatial attention mechanism to improve the contextual reasoning ability of the model. For the second challenge, we propose the theory of multi-scale fusion under the synergistic diffusion mechanism, which effectively solves the problem of insufficient training samples and ensures that the restored part is highly consistent with the original image in terms of style, texture,  and color.

\begin{figure}
  \centering
  \begin{subfigure}{0.49\linewidth}
    \includegraphics[width=\textwidth]{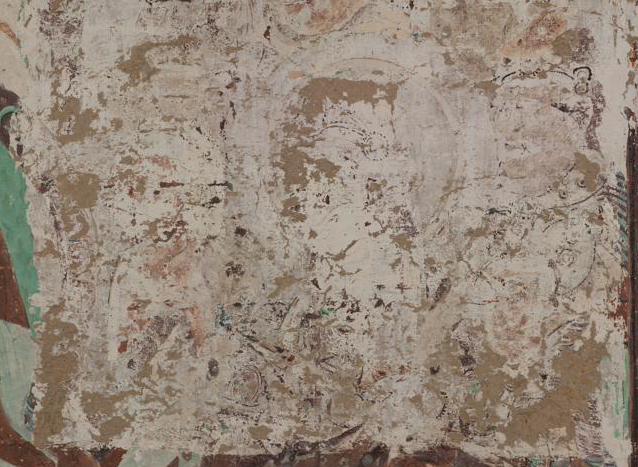}
    \caption{Damaged Image}
    \label{fig: di}
  \end{subfigure}
  \hfill
  \begin{subfigure}{0.49\linewidth}
    \includegraphics[width=\textwidth]{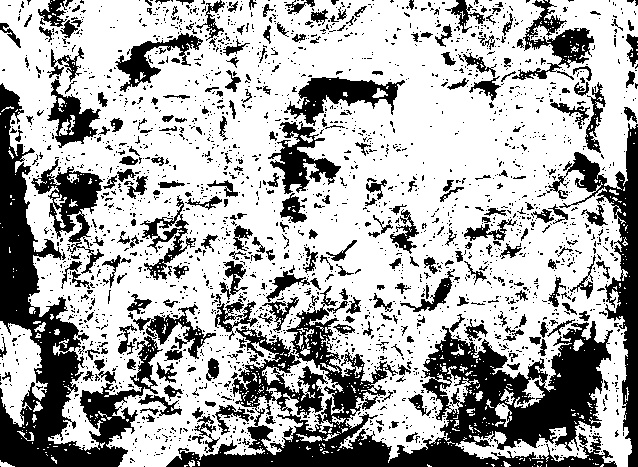}
    \caption{Extracted Contour}
    \label{fig: ec}
  \end{subfigure}
    \caption{Damaged area and its corresponding extracted contour}
  \label{fig: di_ec}
\end{figure}

\subsection{Contour Extraction}
The missing sections of the mural are rarely entirely blank; instead, they typically exhibit faint outline textures, as shown in Fig.~\ref{fig: di}. In the artificial restoration process, these textures serve as valuable references to guide the restorer’s work. Similarly, these textures can also play auxiliary roles in the process of automatic repair.

In this paper, we aim to extract contour information from the missing regions of the image, shown in Fig.~\ref{fig: ec}, and utilize it as conditional guidance to inform the restoration of the damaged areas. Specifically, this paper employs a $K$-Means based approach to implement edge extraction of damaged region images. The damaged image, with dimensions $h \times w $, is first serialized, where each pixel is treated as an individual. This paper applies the $K$-Means clustering algorithm to classify the $hw$ pixels into two categories, with $K=2$. The goal of $K$-means is to minimize the following objective function:
\begin{equation}
J=\sum_{i=1}^n \sum_{k=1}^K r_{i k}\left\|x_i-\mu_k\right\|^2,~  n = hw ,~ K = 2
\end{equation}
where $r_{i k}$ is a binary indicator that equals $1$ if a specified pixel $x_i$ belongs to cluster $k$, and $0$ otherwise. $\mu_k$ is the center of cluster $k$, which is the mean of all pixels assigned to that cluster.
In this case, all pixels are divided into two clusters, foreground and background, allowing for the extraction of the image's contours.

\begin{figure*}[t]
\centering
\includegraphics[width=\linewidth]{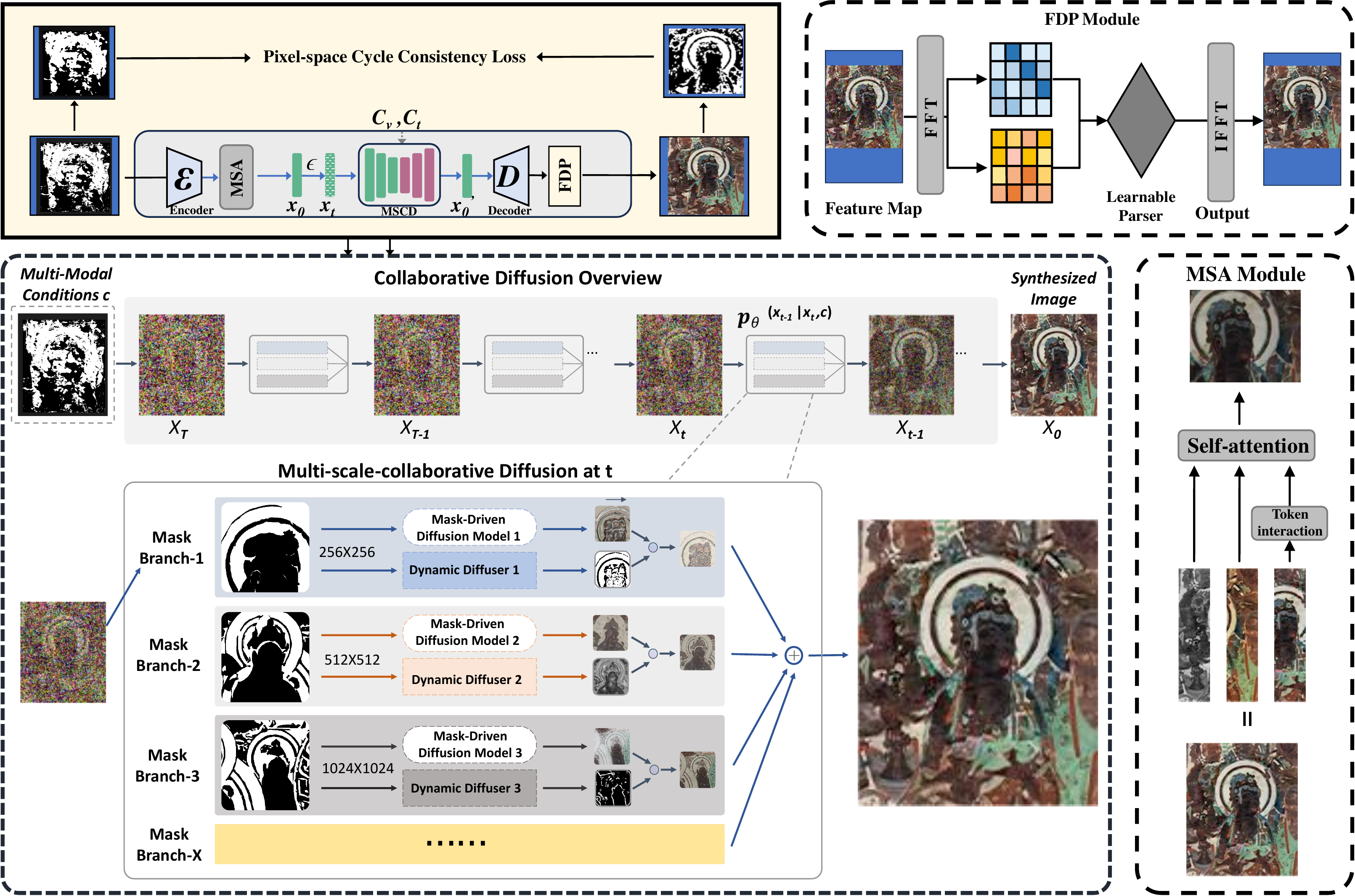}
\caption{An overview of DiffuMural model.}
\label{fig:overview}

\end{figure*}

\subsection{Conditional Controls with Image Generation}
The diffusion model shows powerful performance in image processing, and we introduce the diffusion model as the main architecture for mural restoration.The diffusion model defines a Markovian chain of diffusion forward process $q\left(x_t \mid x_0\right)$ by gradually adding noise to input data $x_0$:
\begin{equation}
x_t=\sqrt{\bar{\alpha}_t} x_0+\sqrt{1-\bar{\alpha}_t} \epsilon, \quad \epsilon \sim \mathcal{N}(\mathbf{0}, I),
\end{equation}
where $\epsilon$ is a noise map sampled, with $\bar{\alpha}_t:=\prod_{s=0}^t \alpha_s$ and $\alpha_t=1-\beta_t$.The diffusion training loss can be represented by:
\begin{equation}
\mathcal{L}\left(\epsilon_\theta\right)=\sum_{t=1}^T \mathbb{E}_{x_0 \sim q\left(x_0\right), \epsilon \sim \mathcal{N}(\mathbf{0}, I)}\left[\left\|\epsilon_\theta\left(\sqrt{\bar{\alpha}_t} x_0+\sqrt{1-\bar{\alpha}_t} \epsilon\right)-\epsilon\right\|_2^2\right] .
\end{equation}
During the inference, given a random noise $x_t\sim \mathcal{N}(\mathbf{0}, I)$, we can predict final denoised image $x_0$ with the step-by-step denoising process:\begin{equation}
x_{t-1}=\frac{1}{\sqrt{\alpha_t}}\left(x_t-\frac{1-\alpha_t}{\sqrt{1-\bar{\alpha}_t}} \epsilon_\theta\left(\mathbf{x}_t, t\right)\right)+\sigma_t \epsilon
\end{equation} In the context of controllable generation, given the image condition $c_v$ and the text prompt $c_t$, the diffusion training loss for time step $t$ can be rewritten as:\begin{equation}
\mathcal{L}_{\text {train }}=\mathbb{E}_{x_0, t, c_t, c_v, \epsilon \sim \mathcal{N}(0,1)}\left[\left\|\epsilon_\theta\left(x_t, t, c_t, c_v\right)-\epsilon\right\|_2^2\right] .
\end{equation}By minimizing the loss of consistency between the input condition $c_v$ and the corresponding output condition $\hat{c}_v$ of the generated image $x_0^{\prime}$, the control of various conditions is optimized in a unified way, and a more controllable generation is realized. The reward consistency loss can be expressed as:\begin{equation}
\begin{aligned}
\mathcal{L}_{\text {reward }} & =\mathcal{L}\left(c_v, \hat{c}_v\right) \\
& =\mathcal{L}\left(c_v, \mathbb{D}\left(x_0^{\prime}\right)\right) \\
& =\mathcal{L}\left(c_v, \mathbb{D}\left[\mathbb{G}^T\left(c_t, c_v, x_T, t\right)\right]\right)
\end{aligned}
\end{equation} $\mathcal{L}$ is an abstract metric function in which $\mathcal{L}$  is the cross-entropy loss per pixel in the context of the use of the segmentation mask as an input condition control in a mural restoration task. Reward model $\mathbb{D}$ also depends on the condition, and we use UperNet as the segmentation mask condition.In addition to the reward loss, we added the diffusion training loss to ensure that the original image generation ability is not affected. Ultimately, the total loss is a combination of$\mathcal{L}_{\text {train}}$ and $\mathcal{L}_{\text {reward }}$:\begin{equation}
\mathcal{L}_{\text {total }}=\mathcal{L}_{\text {train }}+\lambda \cdot \mathcal{L}_{\text {reward }},
\end{equation}

where $\lambda$ is the hyperparameter that adjusts the weight of the reward loss. The consistency loss-guided diffusion model samples at different time steps to obtain images that are consistent with the input control, thereby improving controllability.

\subsection{Murals spatial attention mechanism}

To enhance the model's contextual inference ability and focus on important regions, we integrate a mural-spatial attention mechanism (MSA) into the U-Net framework. This process involves downsampling the input features to expand the receptive field and reduce noise. The downsampled features are transformed into query, key, and value tensors through linear projections. The scaled dot product attention is then applied to compute the attention weights, which are used to aggregate value tensors. Specifically, we use a scaling factor $1 / \sqrt{d_k}$ to prevent vanishing gradients during the dot product computation. Finally, the multi-head self-attention module (MSA) computes attention across multiple subspaces by dividing the query, key, and value tensors into $n$ parts, processing them in parallel, and concatenating the results. This mechanism enables the model to effectively capture interdependencies across input features, improving the overall feature extraction and mural restoration performance.

\subsection{Multi-scale convergence and co-diffusion}
For the task of restoration an mural with a large missing area, we hope that the model has strong context modelling capability as well as richer sensory fields to achieve better repair results. Although multiscale fusion can be a good solution to the problem of global local information connectivity, the difference in the processing of local and global information between context modelling and multiscale fusion mechanisms may lead to the loss of local information or ineffective integration into the global context, and the combination of the two often requires a more complex mechanism to coordinate. Therefore, we propose a scale fusion method based on the synergistic diffusion mechanism, which not only circumvents problems such as incompatible information loss but also can more flexibly assign the weights of different scales of information on the impact of the generated results to achieve better collaborative effects.

At the heart of the co-diffusion mechanism is the dynamic diffuser, which adaptively predicts influence functions to enhance and support multi-scale fusion generation. Given $N$ conditional diffusion models $\left\{\epsilon_{\theta_n}\right\}$ pre-trained by single-scale samples which models the distribution $p\left(\mathbf{x}_0 \mid c_n\right)$.where the modality index $n=1, \ldots, N$, we will sample from $p\left(\mathbf{x}_0 \mid \mathbf{c}\right)$ without changing the pre-trained model,where $\mathbf{c}=$ $\left\{c_1, c_2, \cdots, c_M\right\}$.

In the inverse process of diffusion modelling, noise needs to be predicted at each step, so it must be carefully determined when, where, and how each diffusion model contributes. Corresponding to mural restoration, this means that each step of inferential modelling requires a judgement about which scale of visual information should be referred to, to ensure that each level of image detail is fully exploited and reconstructed.At every diffusion time step $t=T, \ldots, 1$, the influence $\mathbf{I}_{n, t}$ t from every pre-trained diffusion model $\left\{\epsilon_{\theta_n}\right\}$ is adaptively determined by a \textbf{dynamic diffuser}$\mathbf{D}_{\phi_n}$:
\begin{equation}
\mathbf{I}_{n, t}=\mathbf{D}_{\phi_n}\left(\mathbf{x}_t, t, c_n\right),
\end{equation}where $n=1, \ldots, N$ is index of the modalities, $\mathbf{I}_{n, t} \in\mathbb{R}^{h \times w}$, $\mathbf{x}_t$ is the noisy image at time $t, c_n$ is the condition of the $n^{t h}$ modality,where $\mathbf{D}_{\phi_n}$ is the \textbf{dynamic diffuser}implemented by a UNet. To count the overall impact strength, we perform a cross-modal maximum calculation for each pixel, which ultimately results in the impact function $\hat{\mathbf{I}}_{n, t}$:\begin{equation}
\hat{\mathbf{I}}_{n, t, p}=\frac{\exp \left(\mathbf{I}_{n, t, p}\right)}{\sum_{j=1}^N \exp \left(\mathbf{I}_{j, t, p}\right)} .
\end{equation}
We use the learned information function $\hat{\mathbf{I}}_{n, t}$ to control the contribution of each pre-trained diffusion model to accomplish multi-scale fusion and \textbf{collaborative diffusion}:\begin{equation}
\boldsymbol{\epsilon}_{\text {pred }, t}=\sum_{n=1}^N \hat{\mathbf{I}}_{n, t} \odot \boldsymbol{\epsilon}_{\theta_n}\left(x_t, t, c_n\right)
\end{equation} where $\boldsymbol{\epsilon}_{\theta_m}$ is the $m^{\text {th }}$ collaborator, $\odot$ denotes pixel-wise multiplication.

\subsection{Generated image optimization}
To optimize the stability of the generation phase of the model, we need to further optimize the generated results for the problem of blurring in some regions due to the difference in information matching in multi-scale fusion and context modeling. We propose the FDP module to which using the frequency domain to enhance the detailed information of different modalities, including texture and color information. we introduce the learned filter with parameters into the feature space and adjust the convolutional weights to adjust the specific frequencies of some parts of the image, and finally remap the adjusted frequency features back to the explicit space to obtain the final optimized results. the FDP module is essentially an optimizer that fine-tunes the generated results in areas where the visual display is lacking.

\begin{figure*}[t]
\centering
\includegraphics[width=\linewidth]{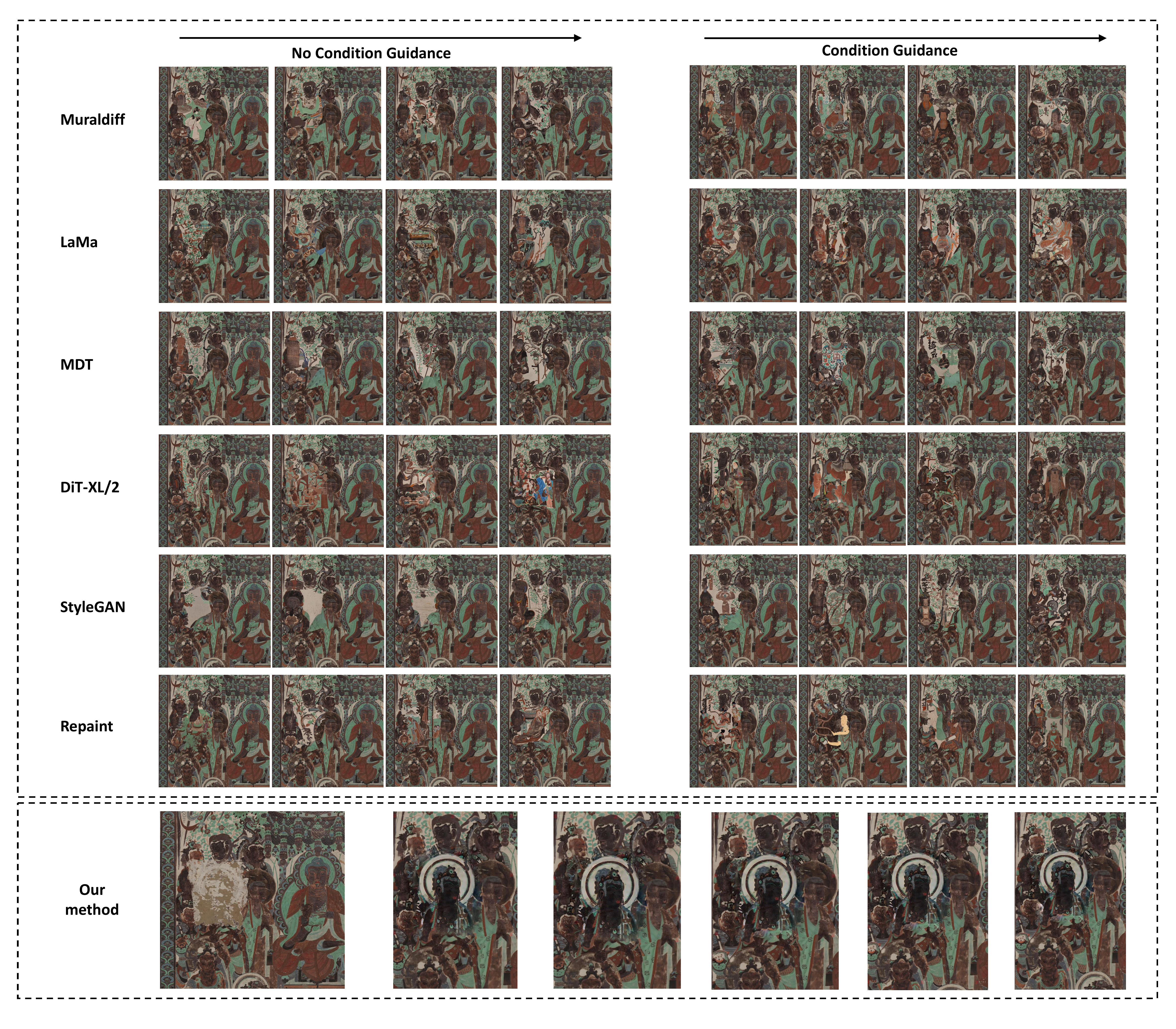}
\caption{The qualitative results of the mural restoration experiments, with each row representing a different restoration method.}
\label{fig:overview}
\end{figure*}

Overall, our method can be summarized in \cref{algo:1}.

\begin{algorithm}
\caption{\textsc{DiffuMural}: Multi-scale Collaborative Mural Restoration}
\label{algo:1}
\KwIn{Damaged mural image $x_0$, contour condition $c_v$, text prompt $c_t$}
\KwOut{Restored mural $\hat{x}_0$}

\textbf{Step 1: Contour Extraction}\;
Extract contour mask $c_v$ from $x_0$ using K-Means clustering with $K=2$\;

\textbf{Step 2: Conditional Guidance Setup}\;
Encode multi-modal conditions $(c_v, c_t)$\;
Prepare multi-scale noisy inputs $x_t^{(n)} \sim \mathcal{N}(0, I)$ at resolutions $\{256, 512, 1024\}$\;

\textbf{Step 3: Multi-scale Collaborative Diffusion}\;
\For{each timestep $t = T \rightarrow 1$}{
    \For{each scale $n \in \{1, 2, 3\}$}{
        Compute influence map $I_{n,t} = D_{\phi_n}(x_t^{(n)}, t, c_n)$\;
    }
    Normalize influence: $\hat{I}_{n,t,p} = \frac{\exp(I_{n,t,p})}{\sum_{j=1}^{N} \exp(I_{j,t,p})}$\;
    Fuse predictions: $\epsilon_{\text{pred},t} = \sum_{n=1}^{N} \hat{I}_{n,t} \odot \epsilon_{\theta_n}(x_t^{(n)}, t, c_n)$\;
    Denoise: $x_{t-1} = \frac{1}{\sqrt{\alpha_t}} \left(x_t - \frac{1-\alpha_t}{\sqrt{1 - \alpha_t}} \epsilon_{\text{pred},t} \right) + \sigma_t \epsilon$\;
}

\textbf{Step 4: Generated Image Optimization}\;
Apply Frequency-Domain Processing (FDP) module to enhance $\hat{x}_0$ for texture and color refinement\;

\Return{$\hat{x}_0$}
\end{algorithm}

\section{Experiments}
In this section, we will conduct experiments to assess the reasonableness and effectiveness of DiffuMural in the mural restoration task.

\textbf{Comparative experiments.} 
We compare DiffuMural with some advanced mural restoration models and recent SOTA methods, including StyleGAN~\cite{karras2019style}, MDT~\cite{gao2023masked}, DiT-XL/2~\cite{peebles2023scalable}, LaMa~\cite{suvorov2022resolution}, RePaint~\cite{lugmayr2022repaint} and Muraldiff~\cite{xu2024muraldiff}. We did not select more specialised mural restoration models for comparison as the mural restoration tasks were different and could not be applied to our dataset.

\textbf{Data.} 
With the assistance of the Dunhuang Academy, we used a laser scanner with a resolution of 0.5$mm$ to collect data from Cave 320 at Dunhuang, getting 27 murals from the Tang dynasty (618-907) with a resolution of 8K, averaging about 40 $m^{2}$ per mural. The murals to be restored are taken out as a test set, and we repeatedly crop the remaining 26 murals according to the scales of 256, 512, and 1024 respectively, with overlapping pixels of $70\%$ of the total pixels, and remove samples containing invalid regions such as black colour carried over from scanning and flaking of the murals themselves. We end up with valid training samples total 420K images.

\textbf{Training Detail.}
All experiments are conducted on an Ubuntu 16.04.1 server equipped with 8 Nvidia A100 GPUs. All codes were developed in Python 3.8.10, PyTorch 1.12.1, and CUDA 11.7 environments. We trained the three mask-driven diffusion models and the dynamic diffuser in DiffuMural using samples with scales of 256, 512, and 10240, respectively, and in particular we iterated the model 1000 times by training it from scratch with a batch size of 64 (8 per GPU). The other models are trained on fixed 512-scale samples, with the training steps remaining consistent. In addition, during the generation process, since this model uses the technical framework of conditionally controllable generation for bootstrap repair, other models were tested twice during the testing phase under conditional and unconditional bootstrapping respectively to compare the results with DiffuMural.

\subsection{Quantitative index}
In the task of mural restoration, traditional metrics like FID cannot be directly applied due to the lack of real-value references. To address this, we propose several quantitative metrics tailored to evaluate restoration quality:

\textbf{structural similarity.}
Structural similarity (SSIM) assesses the similarity in structure, brightness, and contrast between the restored and reference regions. For mural restoration, it is computed by comparing the restored area with undamaged or blurred reference regions:
\begin{equation}
\operatorname{SSIM}(x, y)=\left(\frac{2 \mu_x \mu_y+C_1}{\mu_x^2+\mu_y^2+C_1}\right) \cdot\left(\frac{2 \sigma_{x y}+C_2}{\sigma_x^2+\sigma_y^2+C_2}\right)
\end{equation}
where $x$ and $y$ are a local window of the repaired image and the undamaged image, $\mu_x$ and $\mu_y$ are
the mean of $x$ and $y$ ,representative brightness,$\sigma_x^2$ and $\sigma_y^2$ are the variance of $x$ and $y$, expressing contrast.$\sigma_{x y}$ is the covariance of $x$ and $y$,$C_1$ and $C_2$ are constants.

\textbf{Colour consistency.}
Colour consistency (CCON) evaluates the similarity in color distribution between the restored and surrounding regions. Using color histograms, we calculate the Chi-Square Distance:
\begin{equation}
\chi_{\text{CCON}}^2=\sum_{i=1}^n \frac{\left(H_{\text {repair }}(i)-H_{\text {original }}(i)\right)^2}{H_{\text {repair }}(i)+H_{\text {original }}(i)}
\end{equation}where $H_{\text {repair }}$ and $H_{\text {original }}$ are the colour histograms of the repaired and undamaged regions.$\chi_{\text{CCON}}^2$ is Chi-Square Distance with CCON. The smaller $\chi_{\text{CCON}}^2$ is, the higher the colour consistency

\textbf{Texture consistency.}
Texture consistency (TCON) measures the similarity of texture features between the restored and reference areas. Using Local Binary Patterns (LBP), the LBP histograms of both regions are compared using the same method as color histograms. Higher similarity scores indicate better texture consistency.

\textbf{Edge consistency.}
Edge Consistency (ECON) evaluates the preservation of edge details by comparing the gradient information of the restored and reference edge maps:
\begin{equation}
\text {ECON}=\frac{\sum_{i, j}\left|\nabla E_{\text {repaired }}(i, j)-\nabla E_{\text {original }}(i, j)\right|}{\sum_{i, j}\left|\nabla E_{\text {original }}(i, j)\right|}
\end{equation} 
where $E_{\text {repaired}}$ and $E_{\text {original}}$ are the edge obtained by performing edge detection on the restored image and the original image respectively.

\subsection{Quantitative results}
We compare our method with several state-of-the-art approaches, including StyleGAN~\cite{karras2019style}, MDT~\cite{gao2023masked}, DiT-XL/2~\cite{peebles2023scalable}, LaMa~\cite{suvorov2022resolution}, RePaint~\cite{lugmayr2022repaint}, and Muraldiff~\cite{xu2024muraldiff}. The quantitative results for our dataset are presented in Tables~\ref{tab:task1} and~\ref{tab:task2} for task 1 and task 2, respectively. Our method achieves outstanding results across four different evaluation metrics. Specifically, for both tasks, our method outperforms others on the SSIM and ECON indicators, achieving the best performance. In SSIM, our method improves by 6.33\% and 45.45\%, respectively, compared to the second-best results. Similarly, on the ECON indicator, our method shows an enhancement of 28.24\% and 22.43\% over the second-best methods. Furthermore, for the CCON and TCON metrics, our method performs competitively, with an average gap of only 6.47\% compared to the best results for each metric.

\begin{table}[htbp]
  \centering
  \caption{The models are trained on our dataset and tested on the real damaged mural dataset for Task 1. The best results are highlighted in bold, and the second-best results are marked with \underline{underline}. The same format is applied to the results in Table \ref{tab:task2}.}
  \resizebox{\linewidth}{!}{
    \begin{tabu}{cccccc}
    \tabucline[1.5pt]{-}
    
    \multirow{2}[4]{*}{Model} & \multicolumn{1}{c}{\multirow{2}[4]{*}{\makecell[c]{Condition\\Guidance}}} & \multicolumn{4}{c}{Metrics} \\
\cmidrule{3-6}          &       & SSIM~$\uparrow$  & CCON~$\uparrow$  & TCON~$\uparrow$  & ECON~$\downarrow$ \\
    \midrule
    \multirow{2}[2]{*}{StyleGAN~\cite{karras2019style}} & $\times$      & 0.27  & 0.86  & 0.78  & 15.78 \\
          & \checkmark    & 0.61  & 0.81  & 0.77  & 12.75 \\
    \midrule
    \multirow{2}[2]{*}{MDT~\cite{gao2023masked}} &  $\times$     & 0.55  & 0.89  & 0.24  & 9.27 \\
          & \checkmark    & 0.67  & \textbf{0.97}  & 0.38  & 10.28 \\
    \midrule
    \multirow{2}[2]{*}{DiT-XL/2~\cite{peebles2023scalable}} &  $\times$     & 0.44  & 0.72  & 0.74  & 8.44 \\
          & \checkmark    & 0.69  & 0.77  & 0.86  & 7.72 \\
    \midrule
    \multirow{2}[2]{*}{LaMa~\cite{suvorov2022resolution}} &  $\times$     & 0.31  & 0.84  & 0.62  & 10.02 \\
          & \checkmark    & 0.62  & 0.86  & 0.74  & 12.61 \\
    \midrule
    \multirow{2}[2]{*}{RePaint~\cite{lugmayr2022repaint}} &  $\times$     & 0.42  & 0.89  & 0.54  & 8.41 \\
          & \checkmark    & \underline{0.79}  & 0.87  & 0.65  & 6.27 \\
    \midrule
    \multirow{2}[2]{*}{Muraldiff~\cite{xu2024muraldiff}} &   $\times$    & 0.44  & 0.78  & \textbf{0.91}  & \underline{5.17} \\
          & \checkmark    & 0.76  & 0.88  & 0.77  & 5.42 \\
    \midrule
    Ours  & \checkmark    & \textbf{0.84}  & \underline{0.94}  & \underline{0.86}  & \textbf{3.71} \\
    \tabucline[1.5pt]{-}
    \end{tabu}}%
  \label{tab:task1}%
\end{table}%

\begin{table}[htbp]
  \centering
  \caption{The results on the real damaged mural dataset for task 2, and the performance of these models are trained on our dataset.}
  \resizebox{\linewidth}{!}{
    \begin{tabu}{cccccc}
    \tabucline[1.5pt]{-}
    
    \multirow{2}[4]{*}{Model} & \multicolumn{1}{c}{\multirow{2}[4]{*}{\makecell[c]{Condition\\Guidance}}} & \multicolumn{4}{c}{Metrics} \\
\cmidrule{3-6}          &       & SSIM~$\uparrow$  & CCON~$\uparrow$  & TCON~$\uparrow$  & ECON~$\downarrow$ \\
    \midrule
    \multirow{2}[2]{*}{StyleGAN~\cite{karras2019style}} &  $\times$     & 0.18  & 0.85  & \underline{0.82}  & 12.48 \\
          & \checkmark    & 0.42  & 0.87  & 0.65  & 11.24 \\
    \midrule
    \multirow{2}[2]{*}{MDT~\cite{gao2023masked}} &   $\times$    & 0.39  & 0.92  & 0.44  & 10.21 \\
          & \checkmark    & 0.44  & \textbf{0.94}  & 0.57  & 10.45 \\
    \midrule
    \multirow{2}[2]{*}{DiT-XL/2~\cite{peebles2023scalable}} &  $\times$     & 0.37  & 0.77  & 0.55  & \underline{3.21} \\
          & \checkmark    & 0.49  & 0.76  & 0.68  & 4.48 \\
    \midrule
    \multirow{2}[2]{*}{LaMa~\cite{suvorov2022resolution}} &  $\times$     & 0.18  & 0.85  & 0.51  & 6.52 \\
          & \checkmark    & 0.47  & 0.78  & 0.77  & 6.77 \\
    \midrule
    \multirow{2}[2]{*}{RePaint~\cite{lugmayr2022repaint}} & $\times$      & 0.30  & 0.89  & \textbf{0.92}  & 5.58 \\
          & \checkmark    & \underline{0.55}  & 0.81  & 0.70  & 7.41 \\
    \midrule
    \multirow{2}[2]{*}{Muraldiff~\cite{xu2024muraldiff}} &  $\times$     & 0.27  & \underline{0.92}  & 0.48  & 4.38 \\
          & \checkmark    & 0.42  & 0.87  & 0.65  & 4.15 \\
    \midrule
    Ours  & \checkmark    & \textbf{0.80}  & 0.89  & 0.81  & \textbf{2.49} \\
    \tabucline[1.5pt]{-}
    \end{tabu}}%
  \label{tab:task2}%
\end{table}%

\subsection{Qualitative index}
\begin{figure}[t]
\centering
\includegraphics[width=8cm]{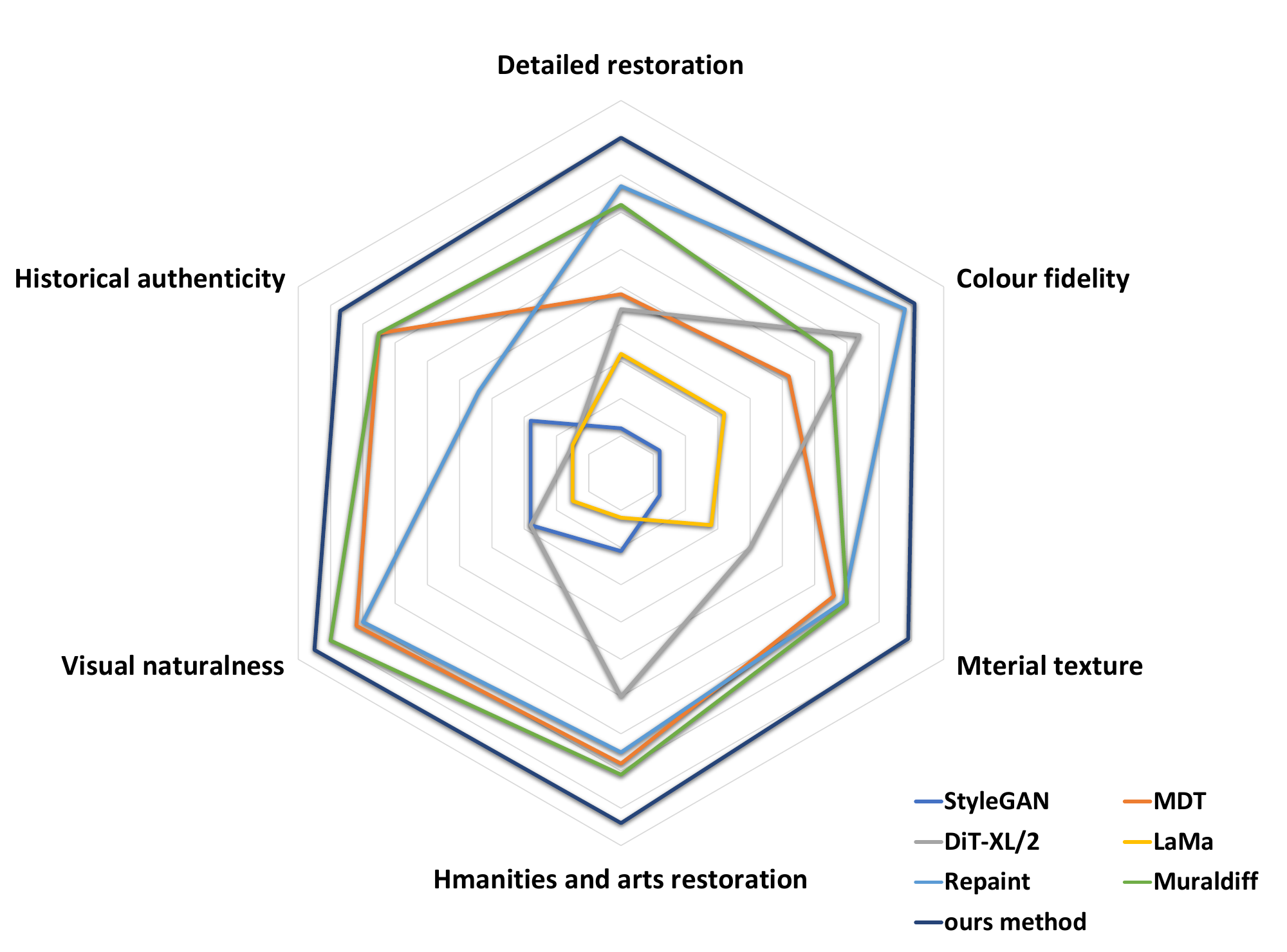}
\caption{Performance of various models in the subjective evaluation system by mural experts.}
\label{fig:qualitative}
\end{figure}

The goal of the evaluation is to ensure that digital mural restoration meets both artistic and technical requirements while preserving cultural significance. Figure \ref{fig:qualitative} illustrates the restoration results evaluated using our proposed qualitative criteria.

In our evaluation, six qualitative metrics are employed to assess the performance of digital mural restoration, covering aspects such as detailed restoration, color fidelity, material texture, humanities and arts restoration, visual naturalness, and historical authenticity. Each metric is scored on a 0-100 scale, with detailed scoring criteria and quality descriptions provided in the Appendix B.

\subsection{Qualitative results}
We conducted comparative restoration experiments on real-world murals. To ensure the professionalism of the restoration process and the objectivity of the evaluation, we construct a human value assessment system. Concretely, we enlisted 289 distinguished mural restoration experts from the Dunhuang Academy, the China Association for the Protection of Cultural Relics, and the Tencent SSV Digital Culture Laboratory. These experts assessed over one hundred restoration outcomes from seven different methods. The evaluations were based on a percentage system across six qualitative indicators, with the final performance of each model determined by the weighted opinions of the experts, as illustrated in Fig.~\ref{fig:qualitative}. The results demonstrate that our proposed DiffuMural excels in restoring damaged areas, particularly in crucial visual metrics such as color consistency, structural integrity, and visual similarity.

Our method performed exceptionally well in the expert evaluation, maintaining a well-balanced set of indicators. This suggests that the proposed approach harmoniously integrates with the cultural and historical significance of the original artwork. While some experts expressed concerns about the ethical and moral dimensions of mural restoration, the method was widely recognized and endorsed by the majority of professionals in the field. Notably, our approach not only represents an innovative solution for restoring large sections of missing murals but also offers a solid foundation for making final restoration decisions in real-world mural conservation efforts.

\section{Discussion: Significance and Social Impact}

The preservation of cultural heritage, particularly ancient murals, is a task of immense historical, artistic, and societal importance. Murals such as those in the Dunhuang Mogao Grottoes are irreplaceable artifacts that embody the spiritual and aesthetic legacy of past civilizations. However, these treasures face increasing risks of irreversible damage due to natural degradation and human activity. Traditional manual restoration methods, while effective, are constrained by time, cost, and ethical concerns. In this context, our proposed \textit{DiffuMural} framework offers a scalable, accurate, and culturally sensitive AI-based solution to aid the digital restoration of such murals.

By incorporating a multi-scale collaborative diffusion mechanism, contour-guided conditioning, and frequency-aware optimization, \textit{DiffuMural} significantly improves the realism, consistency, and controllability of mural restoration. Unlike generic inpainting models, our method is trained on a curated subset of stylistically coherent murals, adhering to restoration ethics by avoiding the hallucination of historically inaccurate content. Furthermore, we introduce a human-centric evaluation system, integrating expert feedback from professional restorers to align the generated results with artistic authenticity and cultural significance.

The broader societal impact of this work lies in its potential to democratize heritage protection. Through responsible AI for Social Good, \textit{DiffuMural} provides museums, researchers, and educators with a powerful tool to virtually restore, preserve, and present damaged cultural assets. This not only safeguards endangered heritage sites but also fosters greater public engagement and cultural appreciation. As a bridge between machine learning and humanistic values, our work represents a step forward in using AI to preserve collective memory for future generations.

\section{Conclusion}
In this study, addressing the challenge of high-resolution mural restoration, particularly the extensive areas of missing murals and the limited availability of samples, we propose an effective generative AI-based solution, named DiffuMural. We depart from the conventional approach of utilizing vast collections of mural data from various dynasties and styles, which is prevalent in current restoration practices. Instead, we adhere to the core principles and ethics of traditional manual restoration by training the model exclusively on a curated collection of 23 murals from the same grotto. On the other hand, we develop a multi-scale collaborative diffusion model, fine-tuning it for exploratory restoration. Specifically, we extract the contours of the damaged areas as inputs to guide the diffusion model for generating inferences, while enhancing feature fusion and co-diffusion mechanisms across different scales. Simultaneously, we introduce quantitative criteria such as stylistic consistency, texture coherence, edge integrity, and structural similarity, establishing a human value assessment system for the restoration outcomes, composed of professional mural restorers. Our results demonstrate that this method achieves superior exploratory restoration results for large-scale missing Dunhuang murals after iterative refinement, offering valuable reference solutions for the manual restoration of murals faced with similar challenges.

\clearpage


\begin{thebibliography}{53}


\ifx \showCODEN    \undefined \def \showCODEN     #1{\unskip}     \fi
\ifx \showISBNx    \undefined \def \showISBNx     #1{\unskip}     \fi
\ifx \showISBNxiii \undefined \def \showISBNxiii  #1{\unskip}     \fi
\ifx \showISSN     \undefined \def \showISSN      #1{\unskip}     \fi
\ifx \showLCCN     \undefined \def \showLCCN      #1{\unskip}     \fi
\ifx \shownote     \undefined \def \shownote      #1{#1}          \fi
\ifx \showarticletitle \undefined \def \showarticletitle #1{#1}   \fi
\ifx \showURL      \undefined \def \showURL       {\relax}        \fi
\providecommand\bibfield[2]{#2}
\providecommand\bibinfo[2]{#2}
\providecommand\natexlab[1]{#1}
\providecommand\showeprint[2][]{arXiv:#2}

\bibitem[Baldassari et~al\mbox{.}(2024)]%
        {baldassari2024conditional}
\bibfield{author}{\bibinfo{person}{Lorenzo Baldassari}, \bibinfo{person}{Ali Siahkoohi}, \bibinfo{person}{Josselin Garnier}, \bibinfo{person}{Knut Solna}, {and} \bibinfo{person}{Maarten~V de Hoop}.} \bibinfo{year}{2024}\natexlab{}.
\newblock \showarticletitle{Conditional score-based diffusion models for Bayesian inference in infinite dimensions}.
\newblock \bibinfo{journal}{\emph{Advances in Neural Information Processing Systems}}  \bibinfo{volume}{36} (\bibinfo{year}{2024}).
\newblock


\bibitem[Batzolis et~al\mbox{.}(2021)]%
        {batzolis2021conditional}
\bibfield{author}{\bibinfo{person}{Georgios Batzolis}, \bibinfo{person}{Jan Stanczuk}, \bibinfo{person}{Carola-Bibiane Sch{\"o}nlieb}, {and} \bibinfo{person}{Christian Etmann}.} \bibinfo{year}{2021}\natexlab{}.
\newblock \showarticletitle{Conditional image generation with score-based diffusion models}.
\newblock \bibinfo{journal}{\emph{arXiv preprint arXiv:2111.13606}} (\bibinfo{year}{2021}).
\newblock


\bibitem[Cao et~al\mbox{.}(2021)]%
        {cao2021superresolution}
\bibfield{author}{\bibinfo{person}{Jianfang Cao}, \bibinfo{person}{Yiming Jia}, \bibinfo{person}{Minmin Yan}, {and} \bibinfo{person}{Xiaodong Tian}.} \bibinfo{year}{2021}\natexlab{}.
\newblock \showarticletitle{Superresolution reconstruction method for ancient murals based on the stable enhanced generative adversarial network}.
\newblock \bibinfo{journal}{\emph{EURASIP Journal on Image and Video Processing}}  \bibinfo{volume}{2021} (\bibinfo{year}{2021}), \bibinfo{pages}{1--23}.
\newblock


\bibitem[Cao et~al\mbox{.}(2020)]%
        {cao2020ancient}
\bibfield{author}{\bibinfo{person}{Jianfang Cao}, \bibinfo{person}{Zibang Zhang}, \bibinfo{person}{Aidi Zhao}, \bibinfo{person}{Hongyan Cui}, {and} \bibinfo{person}{Qi Zhang}.} \bibinfo{year}{2020}\natexlab{}.
\newblock \showarticletitle{Ancient mural restoration based on a modified generative adversarial network}.
\newblock \bibinfo{journal}{\emph{Heritage Science}}  \bibinfo{volume}{8} (\bibinfo{year}{2020}), \bibinfo{pages}{1--14}.
\newblock


\bibitem[Chen et~al\mbox{.}(2019)]%
        {chen2019improving}
\bibfield{author}{\bibinfo{person}{Chen Chen}, \bibinfo{person}{Shuai Mu}, \bibinfo{person}{Wanpeng Xiao}, \bibinfo{person}{Zexiong Ye}, \bibinfo{person}{Liesi Wu}, {and} \bibinfo{person}{Qi Ju}.} \bibinfo{year}{2019}\natexlab{}.
\newblock \showarticletitle{Improving image captioning with conditional generative adversarial nets}. In \bibinfo{booktitle}{\emph{Proceedings of the AAAI conference on artificial intelligence}}, Vol.~\bibinfo{volume}{33}. \bibinfo{pages}{8142--8150}.
\newblock


\bibitem[Chen et~al\mbox{.}(2023)]%
        {chen2023seeing}
\bibfield{author}{\bibinfo{person}{Zijiao Chen}, \bibinfo{person}{Jiaxin Qing}, \bibinfo{person}{Tiange Xiang}, \bibinfo{person}{Wan~Lin Yue}, {and} \bibinfo{person}{Juan~Helen Zhou}.} \bibinfo{year}{2023}\natexlab{}.
\newblock \showarticletitle{Seeing beyond the brain: Conditional diffusion model with sparse masked modeling for vision decoding}. In \bibinfo{booktitle}{\emph{Proceedings of the IEEE/CVF Conference on Computer Vision and Pattern Recognition}}. \bibinfo{pages}{22710--22720}.
\newblock


\bibitem[Chrysos et~al\mbox{.}(2018)]%
        {chrysos2018robust}
\bibfield{author}{\bibinfo{person}{Grigorios~G Chrysos}, \bibinfo{person}{Jean Kossaifi}, {and} \bibinfo{person}{Stefanos Zafeiriou}.} \bibinfo{year}{2018}\natexlab{}.
\newblock \showarticletitle{Robust conditional generative adversarial networks}.
\newblock \bibinfo{journal}{\emph{arXiv preprint arXiv:1805.08657}} (\bibinfo{year}{2018}).
\newblock


\bibitem[Denton et~al\mbox{.}(2016)]%
        {denton2016semi}
\bibfield{author}{\bibinfo{person}{Remi Denton}, \bibinfo{person}{Sam Gross}, {and} \bibinfo{person}{Rob Fergus}.} \bibinfo{year}{2016}\natexlab{}.
\newblock \showarticletitle{Semi-supervised learning with context-conditional generative adversarial networks}.
\newblock \bibinfo{journal}{\emph{arXiv preprint arXiv:1611.06430}} (\bibinfo{year}{2016}).
\newblock


\bibitem[Ding et~al\mbox{.}(2021)]%
        {ding2021ccgan}
\bibfield{author}{\bibinfo{person}{Xin Ding}, \bibinfo{person}{Yongwei Wang}, \bibinfo{person}{Zuheng Xu}, \bibinfo{person}{William~J Welch}, {and} \bibinfo{person}{Z~Jane Wang}.} \bibinfo{year}{2021}\natexlab{}.
\newblock \showarticletitle{Ccgan: Continuous conditional generative adversarial networks for image generation}. In \bibinfo{booktitle}{\emph{International conference on learning representations}}.
\newblock


\bibitem[Dudhane et~al\mbox{.}(2024)]%
        {dudhane2024dynamic}
\bibfield{author}{\bibinfo{person}{Akshay Dudhane}, \bibinfo{person}{Omkar Thawakar}, \bibinfo{person}{Syed~Waqas Zamir}, \bibinfo{person}{Salman Khan}, \bibinfo{person}{Fahad~Shahbaz Khan}, {and} \bibinfo{person}{Ming-Hsuan Yang}.} \bibinfo{year}{2024}\natexlab{}.
\newblock \showarticletitle{Dynamic Pre-training: Towards Efficient and Scalable All-in-One Image Restoration}.
\newblock \bibinfo{journal}{\emph{arXiv preprint arXiv:2404.02154}} (\bibinfo{year}{2024}).
\newblock


\bibitem[Gao et~al\mbox{.}(2023)]%
        {gao2023masked}
\bibfield{author}{\bibinfo{person}{Shanghua Gao}, \bibinfo{person}{Pan Zhou}, \bibinfo{person}{Ming-Ming Cheng}, {and} \bibinfo{person}{Shuicheng Yan}.} \bibinfo{year}{2023}\natexlab{}.
\newblock \showarticletitle{Masked diffusion transformer is a strong image synthesizer}. In \bibinfo{booktitle}{\emph{Proceedings of the IEEE/CVF International Conference on Computer Vision}}. \bibinfo{pages}{23164--23173}.
\newblock


\bibitem[Harvey et~al\mbox{.}(2021)]%
        {harvey2021conditional}
\bibfield{author}{\bibinfo{person}{William Harvey}, \bibinfo{person}{Saeid Naderiparizi}, {and} \bibinfo{person}{Frank Wood}.} \bibinfo{year}{2021}\natexlab{}.
\newblock \showarticletitle{Conditional image generation by conditioning variational auto-encoders}.
\newblock \bibinfo{journal}{\emph{arXiv preprint arXiv:2102.12037}} (\bibinfo{year}{2021}).
\newblock


\bibitem[Huang et~al\mbox{.}(2022)]%
        {huang2022fastdiff}
\bibfield{author}{\bibinfo{person}{Rongjie Huang}, \bibinfo{person}{Max~WY Lam}, \bibinfo{person}{Jun Wang}, \bibinfo{person}{Dan Su}, \bibinfo{person}{Dong Yu}, \bibinfo{person}{Yi Ren}, {and} \bibinfo{person}{Zhou Zhao}.} \bibinfo{year}{2022}\natexlab{}.
\newblock \showarticletitle{Fastdiff: A fast conditional diffusion model for high-quality speech synthesis}.
\newblock \bibinfo{journal}{\emph{arXiv preprint arXiv:2204.09934}} (\bibinfo{year}{2022}).
\newblock


\bibitem[Huang and Hong(2023)]%
        {huang2023diffusion}
\bibfield{author}{\bibinfo{person}{Shaozong Huang} {and} \bibinfo{person}{Lan Hong}.} \bibinfo{year}{2023}\natexlab{}.
\newblock \showarticletitle{Diffusion model for mural image inpainting}. In \bibinfo{booktitle}{\emph{2023 IEEE 7th Information Technology and Mechatronics Engineering Conference (ITOEC)}}, Vol.~\bibinfo{volume}{7}. IEEE, \bibinfo{pages}{886--890}.
\newblock


\bibitem[Isola et~al\mbox{.}(2017)]%
        {isola2017image}
\bibfield{author}{\bibinfo{person}{Phillip Isola}, \bibinfo{person}{Jun-Yan Zhu}, \bibinfo{person}{Tinghui Zhou}, {and} \bibinfo{person}{Alexei~A Efros}.} \bibinfo{year}{2017}\natexlab{}.
\newblock \showarticletitle{Image-to-image translation with conditional adversarial networks}. In \bibinfo{booktitle}{\emph{Proceedings of the IEEE conference on computer vision and pattern recognition}}. \bibinfo{pages}{1125--1134}.
\newblock


\bibitem[Karras et~al\mbox{.}(2019)]%
        {karras2019style}
\bibfield{author}{\bibinfo{person}{Tero Karras}, \bibinfo{person}{Samuli Laine}, {and} \bibinfo{person}{Timo Aila}.} \bibinfo{year}{2019}\natexlab{}.
\newblock \showarticletitle{A style-based generator architecture for generative adversarial networks}. In \bibinfo{booktitle}{\emph{Proceedings of the IEEE/CVF conference on computer vision and pattern recognition}}. \bibinfo{pages}{4401--4410}.
\newblock


\bibitem[Li et~al\mbox{.}(2021)]%
        {li2021restoration}
\bibfield{author}{\bibinfo{person}{Jiao Li}, \bibinfo{person}{Huan Wang}, \bibinfo{person}{Zhiqin Deng}, \bibinfo{person}{Mingtao Pan}, {and} \bibinfo{person}{Honghai Chen}.} \bibinfo{year}{2021}\natexlab{}.
\newblock \showarticletitle{Restoration of non-structural damaged murals in Shenzhen Bao’an based on a generator--discriminator network}.
\newblock \bibinfo{journal}{\emph{Heritage Science}}  \bibinfo{volume}{9} (\bibinfo{year}{2021}), \bibinfo{pages}{1--14}.
\newblock


\bibitem[Li et~al\mbox{.}(2022)]%
        {li2022mat}
\bibfield{author}{\bibinfo{person}{Wenbo Li}, \bibinfo{person}{Zhe Lin}, \bibinfo{person}{Kun Zhou}, \bibinfo{person}{Lu Qi}, \bibinfo{person}{Yi Wang}, {and} \bibinfo{person}{Jiaya Jia}.} \bibinfo{year}{2022}\natexlab{}.
\newblock \showarticletitle{Mat: Mask-aware transformer for large hole image inpainting}. In \bibinfo{booktitle}{\emph{Proceedings of the IEEE/CVF conference on computer vision and pattern recognition}}. \bibinfo{pages}{10758--10768}.
\newblock


\bibitem[Li et~al\mbox{.}(2023)]%
        {li2023lsdir}
\bibfield{author}{\bibinfo{person}{Yawei Li}, \bibinfo{person}{Kai Zhang}, \bibinfo{person}{Jingyun Liang}, \bibinfo{person}{Jiezhang Cao}, \bibinfo{person}{Ce Liu}, \bibinfo{person}{Rui Gong}, \bibinfo{person}{Yulun Zhang}, \bibinfo{person}{Hao Tang}, \bibinfo{person}{Yun Liu}, \bibinfo{person}{Denis Demandolx}, {et~al\mbox{.}}} \bibinfo{year}{2023}\natexlab{}.
\newblock \showarticletitle{Lsdir: A large scale dataset for image restoration}. In \bibinfo{booktitle}{\emph{Proceedings of the IEEE/CVF Conference on Computer Vision and Pattern Recognition}}. \bibinfo{pages}{1775--1787}.
\newblock


\bibitem[Lin et~al\mbox{.}(2018)]%
        {lin2018conditional}
\bibfield{author}{\bibinfo{person}{Jianxin Lin}, \bibinfo{person}{Yingce Xia}, \bibinfo{person}{Tao Qin}, \bibinfo{person}{Zhibo Chen}, {and} \bibinfo{person}{Tie-Yan Liu}.} \bibinfo{year}{2018}\natexlab{}.
\newblock \showarticletitle{Conditional image-to-image translation}. In \bibinfo{booktitle}{\emph{Proceedings of the IEEE conference on computer vision and pattern recognition}}. \bibinfo{pages}{5524--5532}.
\newblock


\bibitem[Liu et~al\mbox{.}(2022)]%
        {liu2022siamtrans}
\bibfield{author}{\bibinfo{person}{Lin Liu}, \bibinfo{person}{Shanxin Yuan}, \bibinfo{person}{Jianzhuang Liu}, \bibinfo{person}{Xin Guo}, \bibinfo{person}{Youliang Yan}, {and} \bibinfo{person}{Qi Tian}.} \bibinfo{year}{2022}\natexlab{}.
\newblock \showarticletitle{Siamtrans: zero-shot multi-frame image restoration with pre-trained siamese transformers}. In \bibinfo{booktitle}{\emph{Proceedings of the AAAI Conference on Artificial Intelligence}}, Vol.~\bibinfo{volume}{36}. \bibinfo{pages}{1747--1755}.
\newblock


\bibitem[Lugmayr et~al\mbox{.}(2022)]%
        {lugmayr2022repaint}
\bibfield{author}{\bibinfo{person}{Andreas Lugmayr}, \bibinfo{person}{Martin Danelljan}, \bibinfo{person}{Andres Romero}, \bibinfo{person}{Fisher Yu}, \bibinfo{person}{Radu Timofte}, {and} \bibinfo{person}{Luc Van~Gool}.} \bibinfo{year}{2022}\natexlab{}.
\newblock \showarticletitle{{RePaint}: Inpainting using denoising diffusion probabilistic models}. In \bibinfo{booktitle}{\emph{Proceedings of the IEEE/CVF conference on computer vision and pattern recognition}}. \bibinfo{pages}{11461--11471}.
\newblock


\bibitem[Luo et~al\mbox{.}(2024)]%
        {luo2024photo}
\bibfield{author}{\bibinfo{person}{Ziwei Luo}, \bibinfo{person}{Fredrik~K Gustafsson}, \bibinfo{person}{Zheng Zhao}, \bibinfo{person}{Jens Sj{\"o}lund}, {and} \bibinfo{person}{Thomas~B Sch{\"o}n}.} \bibinfo{year}{2024}\natexlab{}.
\newblock \showarticletitle{Photo-Realistic Image Restoration in the Wild with Controlled Vision-Language Models}. In \bibinfo{booktitle}{\emph{Proceedings of the IEEE/CVF Conference on Computer Vision and Pattern Recognition}}. \bibinfo{pages}{6641--6651}.
\newblock


\bibitem[Ma et~al\mbox{.}(2022)]%
        {ma2022improved}
\bibfield{author}{\bibinfo{person}{Shang Ma}, \bibinfo{person}{Jianfang Cao}, \bibinfo{person}{Zhaoxia Li}, \bibinfo{person}{Zeyu Chen}, {and} \bibinfo{person}{Xiaohui Hu}.} \bibinfo{year}{2022}\natexlab{}.
\newblock \showarticletitle{An improved algorithm for superresolution reconstruction of ancient murals with a generative adversarial network based on asymmetric pyramid modules}.
\newblock \bibinfo{journal}{\emph{Heritage Science}} \bibinfo{volume}{10}, \bibinfo{number}{1} (\bibinfo{year}{2022}), \bibinfo{pages}{58}.
\newblock


\bibitem[McCormick and Jarman(2005)]%
        {mccormick2005death}
\bibfield{author}{\bibinfo{person}{Jonathan McCormick} {and} \bibinfo{person}{Neil Jarman}.} \bibinfo{year}{2005}\natexlab{}.
\newblock \showarticletitle{Death of a Mural}.
\newblock \bibinfo{journal}{\emph{Journal of Material Culture}} \bibinfo{volume}{10}, \bibinfo{number}{1} (\bibinfo{year}{2005}), \bibinfo{pages}{49--71}.
\newblock


\bibitem[Mirza(2014)]%
        {mirza2014conditional}
\bibfield{author}{\bibinfo{person}{Mehdi Mirza}.} \bibinfo{year}{2014}\natexlab{}.
\newblock \showarticletitle{Conditional generative adversarial nets}.
\newblock \bibinfo{journal}{\emph{arXiv preprint arXiv:1411.1784}} (\bibinfo{year}{2014}).
\newblock


\bibitem[Ni et~al\mbox{.}(2023)]%
        {ni2023conditional}
\bibfield{author}{\bibinfo{person}{Haomiao Ni}, \bibinfo{person}{Changhao Shi}, \bibinfo{person}{Kai Li}, \bibinfo{person}{Sharon~X Huang}, {and} \bibinfo{person}{Martin~Renqiang Min}.} \bibinfo{year}{2023}\natexlab{}.
\newblock \showarticletitle{Conditional image-to-video generation with latent flow diffusion models}. In \bibinfo{booktitle}{\emph{Proceedings of the IEEE/CVF conference on computer vision and pattern recognition}}. \bibinfo{pages}{18444--18455}.
\newblock


\bibitem[Pagnoni et~al\mbox{.}(2018)]%
        {pagnoni2018conditional}
\bibfield{author}{\bibinfo{person}{Artidoro Pagnoni}, \bibinfo{person}{Kevin Liu}, {and} \bibinfo{person}{Shangyan Li}.} \bibinfo{year}{2018}\natexlab{}.
\newblock \showarticletitle{Conditional variational autoencoder for neural machine translation}.
\newblock \bibinfo{journal}{\emph{arXiv preprint arXiv:1812.04405}} (\bibinfo{year}{2018}).
\newblock


\bibitem[Peebles and Xie(2023)]%
        {peebles2023scalable}
\bibfield{author}{\bibinfo{person}{William Peebles} {and} \bibinfo{person}{Saining Xie}.} \bibinfo{year}{2023}\natexlab{}.
\newblock \showarticletitle{Scalable diffusion models with transformers}. In \bibinfo{booktitle}{\emph{Proceedings of the IEEE/CVF International Conference on Computer Vision}}. \bibinfo{pages}{4195--4205}.
\newblock


\bibitem[Qu et~al\mbox{.}(2014)]%
        {qu2014conservation}
\bibfield{author}{\bibinfo{person}{Jianjun Qu}, \bibinfo{person}{Shixiong Cao}, \bibinfo{person}{Guoshuai Li}, \bibinfo{person}{Qinghe Niu}, {and} \bibinfo{person}{Qi Feng}.} \bibinfo{year}{2014}\natexlab{}.
\newblock \showarticletitle{Conservation of natural and cultural heritage in Dunhuang, China}.
\newblock \bibinfo{journal}{\emph{Gondwana Research}} \bibinfo{volume}{26}, \bibinfo{number}{3-4} (\bibinfo{year}{2014}), \bibinfo{pages}{1216--1221}.
\newblock


\bibitem[Radford et~al\mbox{.}(2021)]%
        {radford2021learning}
\bibfield{author}{\bibinfo{person}{Alec Radford}, \bibinfo{person}{Jong~Wook Kim}, \bibinfo{person}{Chris Hallacy}, \bibinfo{person}{Aditya Ramesh}, \bibinfo{person}{Gabriel Goh}, \bibinfo{person}{Sandhini Agarwal}, \bibinfo{person}{Girish Sastry}, \bibinfo{person}{Amanda Askell}, \bibinfo{person}{Pamela Mishkin}, \bibinfo{person}{Jack Clark}, {et~al\mbox{.}}} \bibinfo{year}{2021}\natexlab{}.
\newblock \showarticletitle{Learning transferable visual models from natural language supervision}. In \bibinfo{booktitle}{\emph{International conference on machine learning}}. PMLR, \bibinfo{pages}{8748--8763}.
\newblock


\bibitem[Ren et~al\mbox{.}(2024)]%
        {ren2024dunhuang}
\bibfield{author}{\bibinfo{person}{Hui Ren}, \bibinfo{person}{Ke Sun}, \bibinfo{person}{Fanhua Zhao}, {and} \bibinfo{person}{Xian Zhu}.} \bibinfo{year}{2024}\natexlab{}.
\newblock \showarticletitle{Dunhuang murals image restoration method based on generative adversarial network}.
\newblock \bibinfo{journal}{\emph{Heritage Science}} \bibinfo{volume}{12}, \bibinfo{number}{1} (\bibinfo{year}{2024}), \bibinfo{pages}{39}.
\newblock


\bibitem[Rombach et~al\mbox{.}(2022)]%
        {rombach2022high}
\bibfield{author}{\bibinfo{person}{Robin Rombach}, \bibinfo{person}{Andreas Blattmann}, \bibinfo{person}{Dominik Lorenz}, \bibinfo{person}{Patrick Esser}, {and} \bibinfo{person}{Bj{\"o}rn Ommer}.} \bibinfo{year}{2022}\natexlab{}.
\newblock \showarticletitle{High-resolution image synthesis with latent diffusion models}. In \bibinfo{booktitle}{\emph{Proceedings of the IEEE/CVF conference on computer vision and pattern recognition}}. \bibinfo{pages}{10684--10695}.
\newblock


\bibitem[Shao et~al\mbox{.}(2023)]%
        {shao2023building}
\bibfield{author}{\bibinfo{person}{Huiyang Shao}, \bibinfo{person}{Qianqian Xu}, \bibinfo{person}{Peisong Wen}, \bibinfo{person}{Peifeng Gao}, \bibinfo{person}{Zhiyong Yang}, {and} \bibinfo{person}{Qingming Huang}.} \bibinfo{year}{2023}\natexlab{}.
\newblock \showarticletitle{Building Bridge Across the Time: Disruption and Restoration of Murals In the Wild}. In \bibinfo{booktitle}{\emph{Proceedings of the IEEE/CVF International Conference on Computer Vision}}. \bibinfo{pages}{20259--20269}.
\newblock


\bibitem[Sinha et~al\mbox{.}(2021)]%
        {sinha2021d2c}
\bibfield{author}{\bibinfo{person}{Abhishek Sinha}, \bibinfo{person}{Jiaming Song}, \bibinfo{person}{Chenlin Meng}, {and} \bibinfo{person}{Stefano Ermon}.} \bibinfo{year}{2021}\natexlab{}.
\newblock \showarticletitle{D2c: Diffusion-decoding models for few-shot conditional generation}.
\newblock \bibinfo{journal}{\emph{Advances in Neural Information Processing Systems}}  \bibinfo{volume}{34} (\bibinfo{year}{2021}), \bibinfo{pages}{12533--12548}.
\newblock


\bibitem[Sohn et~al\mbox{.}(2015)]%
        {sohn2015learning}
\bibfield{author}{\bibinfo{person}{Kihyuk Sohn}, \bibinfo{person}{Honglak Lee}, {and} \bibinfo{person}{Xinchen Yan}.} \bibinfo{year}{2015}\natexlab{}.
\newblock \showarticletitle{Learning structured output representation using deep conditional generative models}.
\newblock \bibinfo{journal}{\emph{Advances in neural information processing systems}}  \bibinfo{volume}{28} (\bibinfo{year}{2015}).
\newblock


\bibitem[Suvorov et~al\mbox{.}(2022)]%
        {suvorov2022resolution}
\bibfield{author}{\bibinfo{person}{Roman Suvorov}, \bibinfo{person}{Elizaveta Logacheva}, \bibinfo{person}{Anton Mashikhin}, \bibinfo{person}{Anastasia Remizova}, \bibinfo{person}{Arsenii Ashukha}, \bibinfo{person}{Aleksei Silvestrov}, \bibinfo{person}{Naejin Kong}, \bibinfo{person}{Harshith Goka}, \bibinfo{person}{Kiwoong Park}, {and} \bibinfo{person}{Victor Lempitsky}.} \bibinfo{year}{2022}\natexlab{}.
\newblock \showarticletitle{Resolution-robust large mask inpainting with fourier convolutions}. In \bibinfo{booktitle}{\emph{Proceedings of the IEEE/CVF winter conference on applications of computer vision}}. \bibinfo{pages}{2149--2159}.
\newblock


\bibitem[Wang et~al\mbox{.}(2021)]%
        {wang2021virtual}
\bibfield{author}{\bibinfo{person}{Q Wang}, \bibinfo{person}{M Hou}, {and} \bibinfo{person}{S Lyu}.} \bibinfo{year}{2021}\natexlab{}.
\newblock \showarticletitle{Virtual Restoration of Missing Paint Loss of Mural Based on Generative Adversarial Network}.
\newblock \bibinfo{journal}{\emph{The International Archives of the Photogrammetry, Remote Sensing and Spatial Information Sciences}}  \bibinfo{volume}{46} (\bibinfo{year}{2021}), \bibinfo{pages}{807--811}.
\newblock


\bibitem[Wang et~al\mbox{.}(2024a)]%
        {wang2024methods}
\bibfield{author}{\bibinfo{person}{Shiru Wang}, \bibinfo{person}{Ion Sandu}, \bibinfo{person}{Yulia Ivashko}, \bibinfo{person}{Michal Krupa}, \bibinfo{person}{Anna Krukowiecka-Brzeczek}, \bibinfo{person}{Tetiana Yevdokimova}, \bibinfo{person}{Serhii Stavroyany}, \bibinfo{person}{Oksana Kravchuk}, {and} \bibinfo{person}{Andrei~Victor Sandu}.} \bibinfo{year}{2024}\natexlab{a}.
\newblock \showarticletitle{Methods for the preservation and restoration of Dunhuang wall paintings: foreign experience}.
\newblock \bibinfo{journal}{\emph{International Journal of Conservation Science}} \bibinfo{volume}{15}, \bibinfo{number}{1} (\bibinfo{year}{2024}), \bibinfo{pages}{731--748}.
\newblock


\bibitem[Wang et~al\mbox{.}(2024b)]%
        {wang2024image}
\bibfield{author}{\bibinfo{person}{Siyang Wang}, \bibinfo{person}{Jinghao Zhang}, \bibinfo{person}{Jie Huang}, {and} \bibinfo{person}{Feng Zhao}.} \bibinfo{year}{2024}\natexlab{b}.
\newblock \showarticletitle{Image-free Pre-training for Low-Level Vision}. In \bibinfo{booktitle}{\emph{Proceedings of the 32nd ACM International Conference on Multimedia}}. \bibinfo{pages}{8825--8834}.
\newblock


\bibitem[Wang et~al\mbox{.}(2018)]%
        {wang2018understanding}
\bibfield{author}{\bibinfo{person}{Xiaoguang Wang}, \bibinfo{person}{Ningyuan Song}, \bibinfo{person}{Lu Zhang}, {and} \bibinfo{person}{Yanyu Jiang}.} \bibinfo{year}{2018}\natexlab{}.
\newblock \showarticletitle{Understanding subjects contained in Dunhuang mural images for deep semantic annotation}.
\newblock \bibinfo{journal}{\emph{Journal of documentation}} \bibinfo{volume}{74}, \bibinfo{number}{2} (\bibinfo{year}{2018}), \bibinfo{pages}{333--353}.
\newblock


\bibitem[Wu et~al\mbox{.}(2021)]%
        {wu2021clothgan}
\bibfield{author}{\bibinfo{person}{Qiang Wu}, \bibinfo{person}{Baixue Zhu}, \bibinfo{person}{Binbin Yong}, \bibinfo{person}{Yongqiang Wei}, \bibinfo{person}{Xuetao Jiang}, \bibinfo{person}{Rui Zhou}, {and} \bibinfo{person}{Qingguo Zhou}.} \bibinfo{year}{2021}\natexlab{}.
\newblock \showarticletitle{ClothGAN: generation of fashionable Dunhuang clothes using generative adversarial networks}.
\newblock \bibinfo{journal}{\emph{Connection Science}} \bibinfo{volume}{33}, \bibinfo{number}{2} (\bibinfo{year}{2021}), \bibinfo{pages}{341--358}.
\newblock


\bibitem[Xu et~al\mbox{.}(2024a)]%
        {xu2024boosting}
\bibfield{author}{\bibinfo{person}{Xiaogang Xu}, \bibinfo{person}{Shu Kong}, \bibinfo{person}{Tao Hu}, \bibinfo{person}{Zhe Liu}, {and} \bibinfo{person}{Hujun Bao}.} \bibinfo{year}{2024}\natexlab{a}.
\newblock \showarticletitle{Boosting Image Restoration via Priors from Pre-trained Models}. In \bibinfo{booktitle}{\emph{Proceedings of the IEEE/CVF Conference on Computer Vision and Pattern Recognition}}. \bibinfo{pages}{2900--2909}.
\newblock


\bibitem[Xu et~al\mbox{.}(2024b)]%
        {xu2024comprehensive}
\bibfield{author}{\bibinfo{person}{Zishan Xu}, \bibinfo{person}{Yuqing Yang}, \bibinfo{person}{Qianzhen Fang}, \bibinfo{person}{Wei Chen}, \bibinfo{person}{Tingting Xu}, \bibinfo{person}{Jueting Liu}, {and} \bibinfo{person}{Zehua Wang}.} \bibinfo{year}{2024}\natexlab{b}.
\newblock \showarticletitle{A comprehensive dataset for digital restoration of dunhuang murals}.
\newblock \bibinfo{journal}{\emph{Scientific Data}} \bibinfo{volume}{11}, \bibinfo{number}{1} (\bibinfo{year}{2024}), \bibinfo{pages}{955}.
\newblock


\bibitem[Xu et~al\mbox{.}(2024c)]%
        {xu2024muraldiff}
\bibfield{author}{\bibinfo{person}{Zishan Xu}, \bibinfo{person}{Xiaofeng Zhang}, \bibinfo{person}{Wei Chen}, \bibinfo{person}{Jueting Liu}, \bibinfo{person}{Tingting Xu}, {and} \bibinfo{person}{Zehua Wang}.} \bibinfo{year}{2024}\natexlab{c}.
\newblock \showarticletitle{MuralDiff: Diffusion for Ancient Murals Restoration on Large-Scale Pre-Training}.
\newblock \bibinfo{journal}{\emph{IEEE Transactions on Emerging Topics in Computational Intelligence}} (\bibinfo{year}{2024}).
\newblock


\bibitem[Yan et~al\mbox{.}(2024)]%
        {yan2024image}
\bibfield{author}{\bibinfo{person}{Yao Yan}, \bibinfo{person}{Rui Zhang}, \bibinfo{person}{Hao He}, \bibinfo{person}{Tong Lei}, \bibinfo{person}{Xusheng Zhang}, {and} \bibinfo{person}{Chao Jiang}.} \bibinfo{year}{2024}\natexlab{}.
\newblock \showarticletitle{Image Restoration Technology of Tang Dynasty Tomb Murals Using Adversarial Edge Learning}.
\newblock \bibinfo{journal}{\emph{ACM Journal on Computing and Cultural Heritage}} \bibinfo{volume}{17}, \bibinfo{number}{3} (\bibinfo{year}{2024}), \bibinfo{pages}{1--11}.
\newblock


\bibitem[Zhang et~al\mbox{.}(2023a)]%
        {zhang2023adding}
\bibfield{author}{\bibinfo{person}{Lvmin Zhang}, \bibinfo{person}{Anyi Rao}, {and} \bibinfo{person}{Maneesh Agrawala}.} \bibinfo{year}{2023}\natexlab{a}.
\newblock \showarticletitle{Adding conditional control to text-to-image diffusion models}. In \bibinfo{booktitle}{\emph{Proceedings of the IEEE/CVF International Conference on Computer Vision}}. \bibinfo{pages}{3836--3847}.
\newblock


\bibitem[Zhang et~al\mbox{.}(2025)]%
        {zhang2025fdg}
\bibfield{author}{\bibinfo{person}{Ruicheng Zhang}, \bibinfo{person}{Kanghui Tian}, \bibinfo{person}{Zeyu Zhang}, \bibinfo{person}{Qixiang Liu}, {and} \bibinfo{person}{Zhi Jin}.} \bibinfo{year}{2025}\natexlab{}.
\newblock \showarticletitle{FDG-Diff: Frequency-Domain-Guided Diffusion Framework for Compressed Hazy Image Restoration}.
\newblock \bibinfo{journal}{\emph{arXiv preprint arXiv:2501.12832}} (\bibinfo{year}{2025}).
\newblock


\bibitem[Zhang et~al\mbox{.}(2023b)]%
        {zhang2023image}
\bibfield{author}{\bibinfo{person}{Xiaobo Zhang}, \bibinfo{person}{Donghai Zhai}, \bibinfo{person}{Tianrui Li}, \bibinfo{person}{Yuxin Zhou}, {and} \bibinfo{person}{Yang Lin}.} \bibinfo{year}{2023}\natexlab{b}.
\newblock \showarticletitle{Image inpainting based on deep learning: A review}.
\newblock \bibinfo{journal}{\emph{Information Fusion}}  \bibinfo{volume}{90} (\bibinfo{year}{2023}), \bibinfo{pages}{74--94}.
\newblock


\bibitem[Zhang et~al\mbox{.}(2024)]%
        {zhang2024motion}
\bibfield{author}{\bibinfo{person}{Zeyu Zhang}, \bibinfo{person}{Akide Liu}, \bibinfo{person}{Ian Reid}, \bibinfo{person}{Richard Hartley}, \bibinfo{person}{Bohan Zhuang}, {and} \bibinfo{person}{Hao Tang}.} \bibinfo{year}{2024}\natexlab{}.
\newblock \showarticletitle{Motion mamba: Efficient and long sequence motion generation}. In \bibinfo{booktitle}{\emph{European Conference on Computer Vision}}. Springer, \bibinfo{pages}{265--282}.
\newblock


\bibitem[Zhang et~al\mbox{.}(2018)]%
        {zhang2018decoupled}
\bibfield{author}{\bibinfo{person}{Zhifei Zhang}, \bibinfo{person}{Yang Song}, {and} \bibinfo{person}{Hairong Qi}.} \bibinfo{year}{2018}\natexlab{}.
\newblock \showarticletitle{Decoupled learning for conditional adversarial networks}. In \bibinfo{booktitle}{\emph{2018 IEEE Winter Conference on Applications of Computer Vision (WACV)}}. IEEE, \bibinfo{pages}{700--708}.
\newblock


\bibitem[Zhang et~al\mbox{.}(2023c)]%
        {zhang2023shiftddpms}
\bibfield{author}{\bibinfo{person}{Zijian Zhang}, \bibinfo{person}{Zhou Zhao}, \bibinfo{person}{Jun Yu}, {and} \bibinfo{person}{Qi Tian}.} \bibinfo{year}{2023}\natexlab{c}.
\newblock \showarticletitle{ShiftDDPMs: exploring conditional diffusion models by shifting diffusion trajectories}. In \bibinfo{booktitle}{\emph{Proceedings of the AAAI Conference on Artificial Intelligence}}, Vol.~\bibinfo{volume}{37}. \bibinfo{pages}{3552--3560}.
\newblock


\bibitem[Zhu et~al\mbox{.}(2023)]%
        {zhu2023conditional}
\bibfield{author}{\bibinfo{person}{Yuanzhi Zhu}, \bibinfo{person}{Zhaohai Li}, \bibinfo{person}{Tianwei Wang}, \bibinfo{person}{Mengchao He}, {and} \bibinfo{person}{Cong Yao}.} \bibinfo{year}{2023}\natexlab{}.
\newblock \showarticletitle{Conditional text image generation with diffusion models}. In \bibinfo{booktitle}{\emph{Proceedings of the IEEE/CVF Conference on Computer Vision and Pattern Recognition}}. \bibinfo{pages}{14235--14245}.
\newblock


\end{thebibliography}
\end{document}